\documentclass[letterpaper, 10 pt, conference]{ieeeconf}  

\IEEEoverridecommandlockouts                              

\usepackage[utf8]{inputenc}
\usepackage[T1]{fontenc}
\usepackage{amsmath,amssymb,amsfonts}
\usepackage{graphicx}
\usepackage{textcomp}
\usepackage{xcolor}
\DeclareUnicodeCharacter{0308}{\"{}}
\usepackage{algorithm}
\usepackage{algorithmic}
\usepackage{cite}
\usepackage[breaklinks=true]{hyperref}
\usepackage{xurl}
\usepackage{subcaption}
\usepackage[table]{xcolor}   
\usepackage{colortbl}
\usepackage{booktabs}
\usepackage{multirow}
\newcommand{\compressfootnotetext}{\scriptsize} 

\title{\LARGE \bf
Real-time Spatial-temporal Traversability Assessment via Feature-based Sparse Gaussian Process 
}

\begin{document}

\author{
Zhenyu Hou$^{1\dagger}$, Senming Tan$^{1\dagger}$, Zhihao Zhang$^{1\dagger}$, Long Xu$^{1,2}$, Mengke Zhang$^{1,2}$, Zhaoqi He$^{1}$,\\
Chao Xu$^{1,2}$, Fei Gao$^{1,2}$, and Yanjun Cao$^{1,2}$%
\thanks{\compressfootnotetext
This work was partially supported by the Central Guidance Fund Project for Local Science and Technology Development (Grant~No.\ 2024ZY01015) and the Zhejiang Key Laboratory of Advanced Intelligent Warehousing and Logistics Equipment (Grant~No.\ 2024E10007). Corresponding author: 
 Yanjun Cao, Chao Xu.\protect\\%
$^\dagger$indicates equal contribution(co-first authors). \protect\\%
$^{1}$Huzhou Institute of Zhejiang University, Huzhou 313000, China; $^{2}$State Key Laboratory of Industrial Control Technology, Zhejiang University, Hangzhou 310027, China. E‑mails: \texttt{xiagelearn@gmail.com}, \texttt{yanjunhi@zju.edu.cn}}
}

\maketitle
\thispagestyle{empty}
\pagestyle{empty}

\begin{abstract}


Terrain analysis is critical for the practical application of ground mobile robots in real-world tasks, especially in outdoor unstructured environments. In this paper, we propose a novel spatial-temporal traversability assessment method, which aims to enable autonomous robots to effectively navigate through complex terrains. Our approach utilizes sparse Gaussian processes (SGP) to extract geometric features (curvature, gradient, elevation, etc.) directly from point cloud scans. These features are then used to construct a high-resolution local traversability map. Then, we design a spatial-temporal Bayesian Gaussian kernel (BGK) inference method to dynamically evaluate traversability scores, integrating historical and real-time data while considering factors such as slope, flatness, gradient, and uncertainty metrics. GPU acceleration is applied in the feature extraction step, and the system achieves real-time performance. Extensive simulation experiments across diverse terrain scenarios demonstrate that our method outperforms SOTA approaches in both accuracy and computational efficiency. Additionally, we develop an autonomous navigation framework integrated with the traversability map and validate it with a differential driven vehicle in complex outdoor environments. Our code will be open-source for further research and development by the community,{\url{https://github.com/ZJU-FAST-Lab/FSGP_BGK}}.

\end{abstract}

\section{INTRODUCTION}

Autonomous mobile robots have become essential platforms for environmental perception and intelligent decision-making, revolutionizing operational paradigms in geological exploration, regional security, and ecological monitoring through their advanced terrain navigation capabilities \cite{b1,b2,b3}. Traversability estimation is crucial for autonomous navigation as it informs robots about hazardous regions, thereby reducing the risks associated with navigation\cite{b4}. In recent years, Gaussian process-based traversability analysis has gained significant attention \cite{b5,b6}. The Sparse Gaussian Process (SGP), an enhancement of the traditional Gaussian Process (GP), reduces the computational complexity of GP. Leveraging the continuity of SGP, this approach effectively models uneven terrain for local traversability map generation, facilitating efficient navigation and planning. However, SGP-based methods face two major limitations: Firstly, the accuracy of traversability assessment based on single-frame point clouds remains inadequate; Secondly, these methods fail to integrate historical observation data when evaluating terrain traversability, resulting in high CPU overhead and limited real-time performance due to the computational cost of direct data accumulation.

\begin{figure}[t]
    \centering
    \vspace{0.2cm}
    \includegraphics[width=8.6cm]{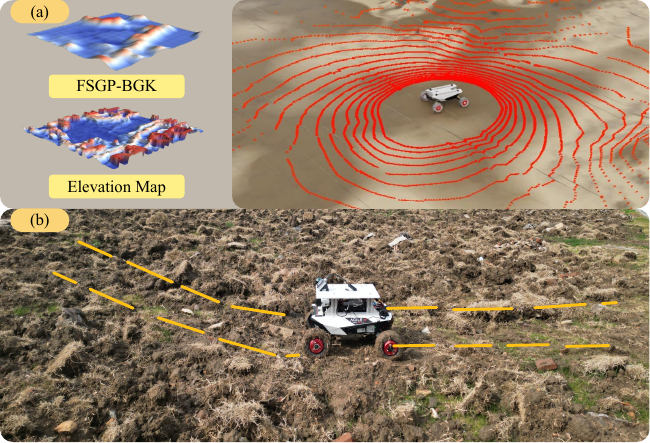}
    \setlength{\abovecaptionskip}{-8pt}
    \caption{(a) Simulation results and the corresponding traversability map; (b) Real-world testing environment, where the yellow line indicates areas that are easy to traverse.}
    \label{fig:first}
    \vspace{-0.7cm}
\end{figure}

To enhance the accuracy of Gaussian process-based traversability assessments, we propose a novel feature-based traversability mapping framework that leverages GPU acceleration for real-time processing. First, local curvature and gradient features are extracted from point cloud data, complemented by feature point extraction and voxel downsampling to create a sparse yet informative representation. Principal Component Analysis (PCA) is then applied to decorrelate these features, ensuring robust and efficient inputs for model training. During the mapping phase, the terrain features are reformulated as a GP regression problem, with an inducing point strategy integrated into the SGP framework. This integration significantly reduces computational complexity while preserving model fidelity, enabling accurate estimation of key parameters such as local curvature, gradient, and slope, thereby facilitating detailed traversability evaluation.

\begin{figure*}[!ht]
  \centering
  \includegraphics[width=1.1\textwidth, trim={1.8cm 1.8cm 2.1cm 0.5cm}, clip]{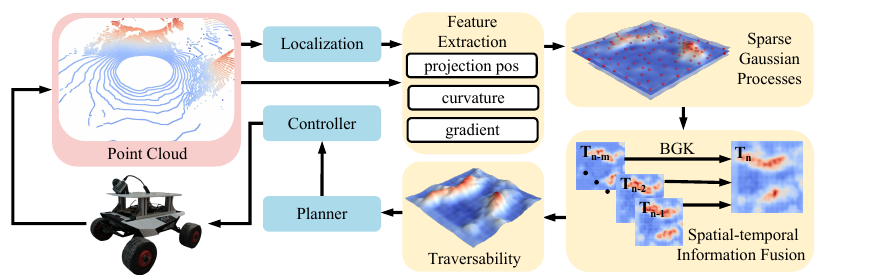}
  \setlength{\abovecaptionskip}{-12pt}
  \caption{Overview of the proposed terrain traversability mapping and navigation framework. From left to right, localization and LiDAR point cloud data are fed into the Feature Extraction module, where curvature and gradient features are computed for each point. These features are then processed by a Sparse Gaussian Process (SGP) model with induced points, yielding local height predictions, variance, and gradient information. Next, a spatial-temporal Bayesian Gaussian Kernel (BGK) fusion step integrates these predictions with historical maps to produce a refined traversability cost map. Finally, we employ A* for trajectory search, MINCO\cite{b7} for trajectory optimization, and a controller for trajectory tracking, thereby generating the necessary control commands for the autonomous vehicle to navigate uneven terrain safely and efficiently.}
  \label{fig:overview}
    \vspace{-0.5cm}   
\end{figure*}

Conventional SGP is unable to integrate historical observational data effectively, as merely accumulating past input data increases computational overhead and reduces processing speed. To address this limitation and better integrate historical observations, we adopt a Bayesian Gaussian Kernel (BGK) method. This approach merges historical data with preliminary traversability estimates. Additionally, Gaussian kernel filtering is applied for local smoothing, resulting in a high-precision, smooth, and robust traversability map. This method effectively resolves the issue of integrating historical data with SGP, while significantly enhancing the system's adaptability and robustness in dynamic environments.

In summary, our main contributions are summarized as follows:
\begin{enumerate}
  \item We present an efficient, feature-driven SGP pipeline for traversability analysis. Our approach seamlessly integrates precise regression with robust uncertainty modeling, thereby enhancing performance in complex environments.
  
  \item We introduce a novel spatial-temporal BGK inference framework to fuse historical observational data into the traversability map. This method significantly improves adaptability in changing environments.

  \item We integrated the traversability assessment function into an autonomous navigation system and open-sourced the framework. The performance has been verified through simulation and real-world experiments in challenging scenarios. Our code will open to foster reproducibility and encourage further research within the community.
\end{enumerate}

\section{RELATED WORK}

\subsection{LiDAR-Based Traversability Assessment }
Recently, there has been an increasing amount of research focused on LiDAR-based traversability map evaluation. In these studies, point cloud data are processed to generate detailed local terrain maps that capture key geometric descriptors such as elevation, curvature, and slope \cite{b8,b9,b10,b11,b12,b13,b14,b15,b16}. Classical methods\cite{b17,b18} extract these features from LiDAR and integrate them with statistical models to construct local and global terrain representations. Although some works\cite{b12}, \cite{b8} extend this approach by incorporating the full state of the $SE(3)$ of the robot or even its suspension dynamics \cite{b16} to improve the fidelity of terrain assessment, such a comprehensive modeling typically imposes a significant computational burden. Consequently, several studies \cite{b10}, \cite{b11}, \cite{b12}, \cite{b13} opt to simplify the problem by restricting the analysis to $\mathbb{R}^2$ space, a compromise that can undermine the accuracy of the risk estimation of traversability. Recent advances in deep learning have prompted the development of self-supervised and end-to-end learning approaches for traversability estimation. J. Seo et al.\cite{b19} proposed a self-supervised framework that takes advantage of vehicle-terrain interaction data to infer traversability directly, thus reducing reliance on manual annotations. Despite their promise, these approaches still face challenges in achieving robust feature extraction in dynamic environments as well as in effectively incorporating historical observational data to maintain global spatial-temporal consistency.

\subsection{Gaussian Process-Based Traversability Assessment }
The GP framework has long been recognized for its effectiveness in modeling continuous spatial phenomena \cite{b18,b20,b21}. To alleviate the computational burden associated with standard GP, SGP methods have been developed, utilizing Bayesian principles to efficiently approximate the full posterior distribution \cite{b22,b23,b24}. In recent years, several works have applied this technique to terrain assessment, aiming to generate smoother and more continuous traversability maps compared to classical elevation maps (EM)\cite{b11}, as shown in the Fig.~\ref{fig:first}. Among the pioneering contributions to uneven terrain traversability analysis, A. Leininger et al. \cite{b25} proposed a SGP-based approach that integrates height, uncertainty, and slope information to enhance path planning. Furthermore, Xue et al. \cite{b26} achieved robust terrain modeling and high-precision traversability analysis by fusing multi-frame LiDAR data, Normal Distributions Transform (NDT) mapping, and spatial-temporal BGK inference. However, these approaches primarily rely on local information, failing to fully utilize historical observation data. This limitation leads to reduced accuracy and poor performance in dynamic environments.

\begin{figure}[h] 
  \centering
   \vspace{-0.3cm}  
  \includegraphics[width=\columnwidth, trim={0.0cm 0.4cm 0.0cm 0.3cm}, clip]{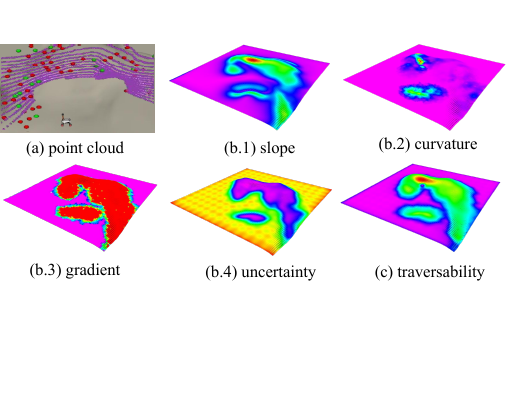}
  \vspace{-2.3cm}   
  \caption{Feature-based SGP results in an uneven environment. From left to right: We begin with the point cloud of the uneven terrain and extract SGP inducing feature points. The SGP then predicts local slope, curvature, gradient, and uncertainty layers, which are finally integrated into a traversability map.}
  \label{fig:feature}
    \vspace{-0.5cm}   
\end{figure}

\section{Traversability Assessment Framework}
In this section, we introduce a novel approach for constructing terrain traversability maps by integrating a terrain-feature-based SGP regression model with a spatial-temporal BGK inference algorithm. GPU acceleration is used to extract position, curvature, and gradient features from point-cloud data, which together form the training set for the SGP model. Meanwhile, an inverse distance weighted interpolation is used to generate a testing set that captures the surrounding terrain's geometric structure. The traversability map is then estimated by fusing the local gradient predicted by the SGP with the local curvature and gradient derived from the testing set. Subsequently, historical map data and the SGP's spatial variance are utilized to refine the map, thereby enhancing accuracy and adaptability. Detailed explanations of each module are provided in the following subsections.

\subsection{Terrain Feature Extraction}
To effectively utilize point cloud data for sparse terrain representation, we extract curvature and gradient features from a point cloud that is aligned with the horizontal axis of the world coordinate system. Let $\mathcal{P}$ denote the original point cloud, where each point $\mathbf{p}_i \in \mathcal{P}$ is represented as $\mathbf{p}_i = (x_i, y_i, z_i)$, with $(x_i, y_i, z_i)$ denoting its three-dimensional spatial coordinates. For each point $\mathbf{p}_i$, the local neighborhood $\mathcal{N}_i$ is determined using the k-nearest neighbors (KNN) algorithm, which selects the $k$ closest points based on Euclidean distance.

\subsubsection{Curvature Computation}
The local curvature is estimated by first computing the centroid $\boldsymbol{\mu}_i = \frac{1}{k} \sum_{\mathbf{p}_j \in \mathcal{N}_i} \mathbf{p}_j$ and the covariance matrix:
\begin{equation}
    \mathbf{C}_i = \frac{1}{k-1} \sum_{\mathbf{p}_j \in \mathcal{N}_i} 
    (\mathbf{p}_j - \boldsymbol{\mu}_i)(\mathbf{p}_j - \boldsymbol{\mu}_i)^\top.
\end{equation}
The curvature is then defined as $\kappa_i = \lambda_{\min} / (\sum_j \lambda_j + \epsilon)$, where $\lambda_{\min}$ is the smallest eigenvalue of $\mathbf{C}_i$, and $\epsilon$ is a small constant to prevent division by zero.

\subsubsection{Gradient Computation}
The local gradient quantifies elevation variation and is given by $g_i = \frac{1}{k} \sum_{\mathbf{p}_j \in \mathcal{N}_i} \lvert z_j - z_i \rvert$, where $z_j$ and $z_i$ are the elevation values of $\mathbf{p}_j$ and $\mathbf{p}_i$, respectively.

\subsubsection{Feature Point Identification}
Feature points are identified using predefined curvature and gradient thresholds $\tau_{\kappa}$ and $\tau_g$. A point is classified as a feature point if $\kappa_i > \tau_{\kappa}$ or $g_i > \tau_g$, forming the feature set $\mathcal{F} = \{\mathbf{p}_i \mid \kappa_i > \tau_{\kappa} \text{ or } g_i > \tau_g\}$. To reduce redundancy, uniform downsampling is applied to $\mathcal{P} - \mathcal{F}$ by partitioning the space into voxel grids of size $v$, retaining one point per voxel to form the downsampled set $\mathcal{D}$.

\subsubsection{Final Point Cloud Assembly}
The final point cloud dataset is given by $\mathcal{P}_{\text{final}} = \mathcal{F} \cup \mathcal{D}$. If the total number of retained points exceeds a threshold $M$, a random sampling strategy is applied:
\begin{equation}
    \mathcal{P}_{\text{final}} = \{(x_i, y_i, z_i, \kappa_i, g_i)\}_{i=1}^{M} 
    = \text{RandomSample}(\mathcal{F} \cup \mathcal{D}, M).
\end{equation}

\subsubsection{Feature Decorrelation and Acceleration}
After extraction, PCA reduces redundancy by projecting data onto the principal axes of maximal variance, improving model efficiency and stability. Feature points with high curvature or large gradients—indicative of critical terrain structures such as ridges, valleys, and cliffs—are prioritized. GPU acceleration is employed for KNN search, covariance computation, eigenvalue decomposition, and PCA, significantly improving computational efficiency.

\subsection{Terrain Feature SGP Model}

We employ a SGP model for terrain representation. An inducing point set $\mathbf{Z}$, corresponding to $\mathcal{P}_{\text{final}}$, is introduced for computational efficiency. The training set is defined as $\mathcal{D}_{\text{train}} = \{(\mathbf{X}_i, z_i)\}$, where $\mathbf{X}_i = (x_i, y_i, \kappa_i, g_i)$ are decorrelated features, and $z_i$ is the terrain elevation. We assume a GP prior:
\begin{equation}
    f(\mathbf{X}) \sim GP\bigl(m(\mathbf{X}),\,k(\mathbf{X}, \mathbf{X}')\bigr),
\end{equation}
where $m(\mathbf{X})$ is the mean and $k(\mathbf{X}, \mathbf{X}')$ is the kernel function.

To construct $\mathcal{D}_{\text{test}}$, we generate test points $\mathcal{G}$ by partitioning the space into a uniform grid. For each $\mathbf{X}^* = (x^*, y^*)$, the $K$ nearest neighbors are identified via KNN. The local curvature and gradient are interpolated by
\begin{IEEEeqnarray}{rCl}
    \kappa^* &=& \sum_{k=1}^{K} w_k\,\kappa_k, \quad g^* \;=\; \sum_{k=1}^{K} w_k\,g_k, \\
    w_k &=& \frac{1}{d_k + \epsilon}, \quad \sum_{k=1}^{K} w_k \;=\; 1,
\end{IEEEeqnarray}
yielding $\mathbf{X}^* = (x^*, y^*, \kappa^*, g^*)$, which is then fed into the trained GP model to predict elevation.

For the test set $\mathbf{X}_*$, the predictive mean and variance are computed as
\begin{IEEEeqnarray}{rCl}
    f^* &=& \mathbf{K}_{*M}\,\mathbf{K}_{MM}^{-1}\,\mathbf{z}_M, \\
    \sigma_*^2 &=& k_{**} \;-\; \mathbf{K}_{*M}\,\mathbf{K}_{MM}^{-1}\,\mathbf{K}_{M*},
\end{IEEEeqnarray}
where $\mathbf{K}_{*M}$ and $\mathbf{K}_{MM}$ are kernel matrices between test and inducing points, and $\mathbf{z}_M$ contains target values associated with $\mathbf{Z}$. This approximation reduces computational complexity while maintaining high accuracy.

\subsection{Traversability Map Construction}

We construct a traversability cost map, \(M_\tau\), by leveraging local features predicted by our SGP and BGK inference methods. First, we compute preliminary traversability estimates from curvature, gradient, and slope information derived from the SGP. We then refine these estimates by fusing historical data via the BGK method and subsequently apply Gaussian kernel filtering to ensure spatial coherence.

For each test point \(\mathbf{X}^* = (x, y, \kappa^*, g^*)\), where \(\kappa^*\) and \(g^*\) denote the curvature and local height gradient predicted by the SGP, and \(\left|\nabla f_*\right|\) represents the slope magnitude, we compute a preliminary traversability score as
\begin{equation}
    \label{eq:pre_trav}
    M_{\tau,\mathrm{pre}} = w_{\kappa}\,\kappa^* + w_g\,g^* + w_{\mathrm{grad}}\,\bigl|\nabla f_*\bigr|,
\end{equation}
where the weights satisfy \(w_{\kappa} + w_g + w_{\mathrm{grad}} = 1\). These weights can be tuned experimentally or learned from data to suit specific application needs.

To improve spatial-temporal consistency, we fuse the preliminary estimates with historical observations stored for each grid cell \((x,y)\). Each cell maintains its prior traversability estimate \(M_{\tau,t-1}\), variance \(\sigma_{\tau,t-1}^2\), and timestamp \(t_0\). We define a temporal decay weight \(\omega_t = \exp\bigl(-\lambda (t - t_0)\bigr)\) and an uncertainty-based confidence weight \(\omega_\sigma = 1/(\sigma_{\tau,t-1}^2 + \epsilon)\). The BGK fusion update is 
\begin{equation}
    \label{eq:bgk_update}
    M_\tau = \frac{\omega_t\,\omega_\sigma\,M_{\tau,t-1} + M_{\tau,\mathrm{pre}}}{\omega_t\,\omega_\sigma + 1},
\end{equation}
with the corresponding variance updated as
\begin{equation}
    \label{eq:bgk_variance}
    \sigma_{\tau,t}^2 = \frac{\omega_t\,\sigma_{\tau,t-1}^2 + \sigma_{\mathrm{pre}}^2}{\omega_t + 1}.
\end{equation}

Finally, to reduce local discontinuities, we smooth \(M_\tau\) via Gaussian filtering. For each grid cell \(\mathbf{X}_i\), the smoothed traversability is computed as
\begin{equation}
    \label{eq:Mtau_smooth}
    M_{\tau,\mathrm{smooth}} = \sum_{j \in \mathcal{N}(i)} k(\mathbf{X}_i,\mathbf{X}_j) \, M_{\tau,j},
\end{equation}
where the Gaussian kernel is defined by \(k(\mathbf{X}_i,\mathbf{X}_j) = \exp\!\bigl(-\|\mathbf{X}_i - \mathbf{X}_j\|^2 / (2\,\sigma^2)\bigr)\). 
The complete algorithmic is outlined in Algorithm~1.

\begin{algorithm}
\caption{Traversability Map Algorithm}
\begin{algorithmic}[1]
\STATE \textbf{Input:} \(\mathbf{X}^*,\, f^*,\, \sigma_*^2\)
\STATE \textbf{Output:} \(M_{\tau,\mathrm{smooth}}\)

\STATE \(M_{\tau,\mathrm{pre}} \leftarrow \text{initializeTraversability}(\mathbf{X}^*,\, f^*,\, \sigma_*^2)\)
\STATE \(\text{mapHistoryBuffer} \leftarrow \text{addValue}(M_{\tau,\mathrm{pre}},\, \mathbf{X}^*,\, \sigma_*^2)\)

\FOR{\textbf{each} \(M_{\tau,i}\,\text{in}\,\text{mapHistoryBuffer}\)}
\STATE \(M_{\tau,i} \leftarrow \text{updateTraversabilityHistory}(M_{\tau,\mathrm{pre}})\)
\ENDFOR

\FOR{\textbf{each} \(\sigma_{\tau,i}^2\,\text{in}\,\text{mapHistoryBuffer}\)}
\STATE \(\sigma_{\tau,i}^2 \leftarrow \text{updateUncertaintyHistory}(\sigma_*^2)\)
\ENDFOR

\FOR{\textbf{each} \(\mathbf{X}_i\,\text{in}\,\text{mapHistoryBuffer}\)}
\STATE \(\mathcal{N}(i) \leftarrow \text{computeNeighboringPoints}(\mathbf{X}_i)\)
\FOR{\textbf{each} \(\bigl(\mathbf{X}_j,\, M_{\tau,j}\bigr)\,\text{in}\,\mathcal{N}(i)\)}
\STATE \(M_{\tau,j} \leftarrow \text{smoothData}(\mathbf{X}_i,\, \mathbf{X}_j,\, M_{\tau,j})\)
\ENDFOR
\ENDFOR

\STATE \(M_{\tau,\mathrm{smooth}} \leftarrow \text{getLatestTraversability}(\text{mapHistoryBuffer})\)

\end{algorithmic}
\end{algorithm}

As shown in Algorithm 1, the traversability smoothing procedure accepts test points \(\mathbf{X}^*\), predicted values \(f^*\) (derived from curvature and gradient), and uncertainties \(\sigma_*^2\) as inputs to produce a smoothed traversability map \(M_{\tau,\mathrm{smooth}}\). The algorithm initializes a preliminary traversability map, leverages a sliding window to maintain historical traversability and uncertainty data, and applies spatial smoothing by adjusting values based on adjacent points points. The refined traversability map is subsequently extracted from the sliding window, encapsulating an enhanced representation of terrain navigability. This approach yields a high-precision, spatial-temporally consistent traversability map that robustly supports autonomous navigation and path planning.

\subsection{System Framework and Implementation}
\begin{figure}[ht]
  \centering
  \includegraphics[width=\columnwidth, trim={0.0cm 0.0cm 0.0cm 0.0cm}, clip]{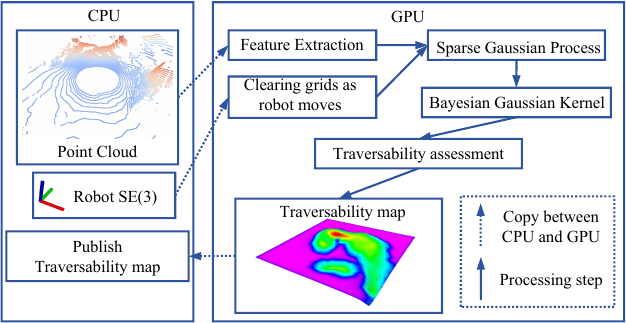}
  \caption{The process of operations on the CPU and GPU during traversability map generation.}
  \label{fig:process}
\end{figure}

As illustrated in Fig.~\ref{fig:overview}, our terrain traversability mapping and navigation framework comprises four main components: localization, mapping, planning, and control. We use a LiDAR-based approach to estimate the robot’s SE(3) pose, while simultaneously feeding raw point cloud data into a feature extraction module that computes curvature and gradient information. These features are then processed by a SGP to predict local terrain attributes, and the resulting predictions are fused with historical observations via a BGK method to produce the traversability map.As Fig.~\ref{fig:process} shown,the entire process is accelerated using a GPU.

To validate our framework, we employ A* for global path planning and use MINCO~\cite{b6} and DDR-opt~\cite{b27} for trajectory optimization and smoothing based on the generated traversability map. Model predictive control (MPC) is then applied for real-time trajectory tracking. We deploy this system on a physical robot, updating the traversability map online and enabling robust, autonomous navigation over uneven terrain in both simulation and real-world experiments.

\section{Experimental Evaluation}
To comprehensively evaluate the proposed FSGP-BGK method, we conducted a series of experiments comprising both quantitative assessments and real-world tests. The quantitative evaluation benchmarks FSGP-BGK against well-established methods, such as SGP based terrain assessment pipeline\cite{b25} and elevation map (EM)\cite{b11}, across various terrain types and different inducing point numbers. Meanwhile, we validated the applicability of the method on physical robotic platforms, with a focus on real-time mapping performance and collision avoidance. 

\subsection{Data Sample and Simulation Setup}

\subsubsection{Terrain Generation and Data Sample}
To rigorously evaluate the performance of the proposed method across a range of challenging scenarios, we utilized the EPFL terrain generator\footnote{\url{https://github.com/droduit/procedural-terrain-generation}} to synthesize large-scale terrain point clouds. These terrains span a 50 m $\times$ 50 m area and encompass four distinct types: In our experiments, we simulated five types of terrain: Hilly Terrain, an environment characterized by small hills and open stretches generated using the Procedural Terrain Generator; Forest Terrain, created by overlaying randomly distributed, unstructured trees of various sizes on the hilly landscape; Ruin Terrain, enhanced by adding both unstructured trees and structured rectangular and cylindrical objects atop the hilly terrain; Urban Terrain, constructed by segmenting the Complex Urban Dataset\cite{b28} into square sections to mimic realistic urban on-road settings; and an Indoor Scene, built using the SceneNet dataset\footnote{\url{https://bitbucket.org/robotvault/downloadscenenet/src/master/}} and segmented into predefined sections to simulate realistic indoor environments.

\begin{figure}[ht] 
  \centering
  \includegraphics[width=\columnwidth, trim={0.0cm 4.0cm 0.0cm 0.0cm}, clip]{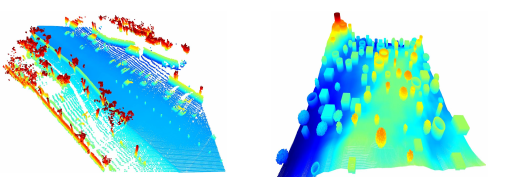}
  \caption{Left: Urban Terrain; Right: Ruin Terrain. Other terrain types are discussed later.}
  \label{fig:small}
\end{figure}

For each terrain type, we generate 500 scenarios with different point clouds, yielding a total of 2500 samples. These point clouds are processed using a specified formula (denoted as $[x]$) to create ground truth traversability maps. These maps are then used to guide A* path planning, followed by trajectory optimization using the MINCO algorithm \cite{b6}, simulating robotic navigation tasks.

\subsubsection{Simulation Configuration}
In our simulations, the robot navigates each terrain type with 15\% of the point cloud data intentionally occluded, thereby mimicking real-world LiDAR limitations such as shadowed regions or sparse sensor scans. The acquired point clouds are processed sequentially and fed into the models at fixed intervals. Traversability estimates are then evaluated by computing the mean error and variance relative to the ground truth—that is, the average difference between the generated traversability map and the ground truth global map. Note that both the mean error and variance are dimensionless and range from 0 to 1, as they represent the traversability scores.

To ensure a fair comparison with EM\cite{b10}, we exclude the gap regions caused by occlusions and focus solely on areas with available data. All simulations are run on a dedicated workstation equipped with an Intel i5-12400F CPU, 16 GB of RAM, and an NVIDIA RTX 4060 GPU, ensuring consistency in computational conditions across all trials.

\subsubsection{Comparative Baselines}
We benchmark FSGP-BGK against the following methods:
\begin{itemize}
    \item SGP: A classical SGP based traversability analysis for terrain mapless navigation\cite{b25}.
    \item FSGP: A feature-based SGP approach proposed in this study, which does not incorporate BGK..
    \item EM: A widely adopted grid-based spatial representation method\cite{b11}, valued for simplicity but prone to gaps in unobservable regions.
\end{itemize}

\subsection{Quantitative Assessment}

\subsubsection{Performance across Diverse Terrains}
\begin{figure}[ht]
  \centering
  \includegraphics[width=\columnwidth, trim={0.0cm 0.27cm 0.0cm 0.0cm}, clip]{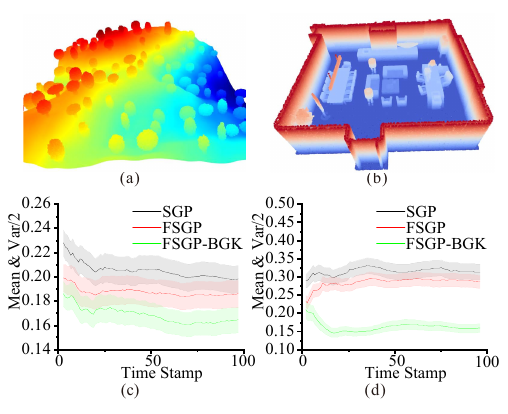}
  \caption{(a) Forest scene point cloud rendered using the EPFL terrain generator; 
  (b) Indoor scene point cloud extracted from an open-source dataset\protect\footnote[4]; 
  (c) and (d) illustrate the temporal evolution of the mean error and variance.}
  \label{fig:Compare}
    \vspace{-0.3cm}   
\end{figure}

As illustrated in Figs.~\ref{fig:Compare}, we randomly select a point cloud from both an outdoor forest environment and an indoor room environment. We then apply three methods, SGP (baseline), FSGP and FSGP-BGK to generate traversability estimates for each trajectory point in chronological order. The mean error and variance, computed relative to the ground truth, are plotted over time in Fig.~\ref{fig:Compare}.

Our experimental results demonstrate that in the outdoor forest scenario with 125 inducing points, FSGP-BGK improves mean accuracy by 18\% to 20\% compared to the baseline SGP and exhibits significantly lower error variance. In the indoor environment, FSGP-BGK outperforms SGP by up to 46\% to 48\% on average while maintaining superior stability. Moreover, in both scenarios the mean error steadily decreases over time, highlighting the enhanced terrain estimation accuracy achieved through the integration of historical observations.

\subsubsection{Algorithmic Scalability in Large-Scale Testing}
To assess the robustness and scalability of our approach, we conducted experiments on 2500 terrain point clouds across five categories: Hilly, Forest, Ruin, Road, and Indoor. Table~\ref{tab:experiment_results} reports the average mean error and variance for each method (lower values indicate better performance, with the best results highlighted in blue). The Indoor dataset and the Road dataset are consistent with those used earlier, while the Hilly, Forest, and Ruin terrains were generated using the EPFL terrain generator. Our FSGP-BGK method consistently achieves the lowest mean error and variance, outperforming both the baseline SGP and the intermediate FSGP variant.

For example, in the Hilly environment, FSGP-BGK reduces the mean error from 0.2604 (SGP) to 0.1237 and the variance from 0.0392 to 0.0085, corresponding to improvements of approximately 52.5\% and 78.3\%, respectively. Similar gains are observed in the Forest and Ruin terrains, where our method not only lowers error but also stabilizes uncertainty estimates. In the Indoor scenario, although FSGP alone yields a slightly higher mean error than SGP (0.2537 vs.\ 0.2455), incorporating BGK reduces it significantly to 0.1978.

These findings underscore two principal insights. First, accurate traversability prediction fundamentally relies on extracting high-density terrain features—such as curvature and gradient—which serve as precise proxies for underlying navigability. Second, by integrating historical observations through our spatial-temporal BGK framework, the method effectively captures temporal dynamics and progressively reduces prediction errors. This dual strategy not only outperforms conventional SGP approaches but also deepens our understanding of the critical determinants of terrain traversability.

\begin{table}[t]
    \centering
    \caption{%
        Experimental Results Across Different Terrains. 
    }
    \label{tab:experiment_results}
    \begin{tabular}{llcc}
    \toprule
    \textbf{Type} & \textbf{Method} & \textbf{Average Mean} & \textbf{Average Variance} \\ 
    \midrule
    \multirow{3}{*}{Hilly} 
        & SGP                     & 0.2604 & 0.0392 \\ 
        & FSGP                  & 0.2321 & 0.0345 \\ 
        & \cellcolor{blue!20}\textbf{FSGP-BGK} & \cellcolor{blue!20}\textbf{0.1237} & \cellcolor{blue!20}\textbf{0.0085} \\ 
    \midrule
    \multirow{3}{*}{Forest} 
        & SGP                     & 0.1831 & 0.0245 \\ 
        & FSGP                  & 0.1777 & 0.0227 \\ 
        & \cellcolor{blue!20}\textbf{FSGP-BGK} & \cellcolor{blue!20}\textbf{0.1687} & \cellcolor{blue!20}\textbf{0.0213} \\ 
    \midrule
    \multirow{3}{*}{Ruin} 
        & SGP                     & 0.2003 & 0.0239 \\ 
        & FSGP                  & 0.1965 & 0.0219 \\ 
        & \cellcolor{blue!20}\textbf{FSGP-BGK} & \cellcolor{blue!20}\textbf{0.1831} & \cellcolor{blue!20}\textbf{0.0206} \\ 
    \midrule
    \multirow{3}{*}{Road} 
        & SGP                     & 0.1694 & 0.0373 \\ 
        & FSGP                  & 0.1434 & 0.0328 \\ 
        & \cellcolor{blue!20}\textbf{FSGP-BGK} & \cellcolor{blue!20}\textbf{0.1051} & \cellcolor{blue!20}\textbf{0.0240} \\ 
    \midrule
    \multirow{3}{*}{Indoor} 
        & SGP                     & 0.2455 & 0.0451 \\ 
        & FSGP                  & 0.2537 & 0.0429 \\ 
        & \cellcolor{blue!20}\textbf{FSGP-BGK} & \cellcolor{blue!20}\textbf{0.1978} & \cellcolor{blue!20}\textbf{0.0420} \\ 
    \bottomrule
    \end{tabular}
\end{table}


\subsubsection{Impact of Inducing Points}
\begin{figure}[h]
  \vspace{-0.4cm}   

  \centering
  \includegraphics[width=\columnwidth, trim={0.0cm 0.5cm 0.0cm 0.0cm}, clip]{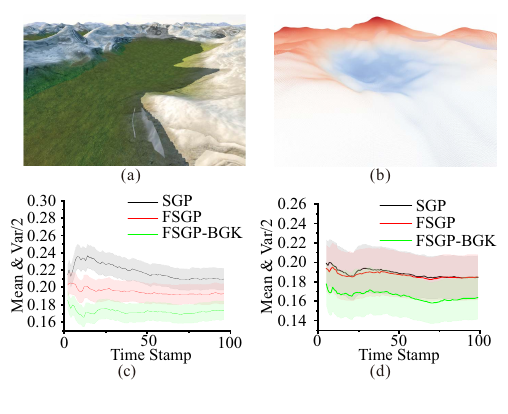}
  \caption{
    (a) Uneven Hilly terrain generated by EPFL terrain generator; 
    (b) Global point cloud extracted from the Hilly terrain; 
    (c) Temporal evolution of the mean error and variance for SGP, FSGP, and FSGP-BGK with 50 inducing points relative to the ground truth; 
    (d) Temporal evolution of the mean error and variance for the same methods with 500 inducing points. 
    This figure visually demonstrates the differences in modeling accuracy and stability among the methods under varying numbers of inducing points.
  }
  \label{fig:number}
  \vspace{-0.2cm}   
\end{figure}

As illustrated in Figs.~\ref{fig:number}(a-b), we generate an uneven Hilly terrain and employ our method to construct the corresponding global point cloud. We then investigate the influence of Inducing points by evaluating SGP, FSGP, and FSGP-BGK with both 50 and 500 inducing points, as shown in Figs.~\ref{fig:number}(c-d). When only 50 inducing points are used, FSGP-BGK outperforms the baseline SGP by up to 19\% in mean accuracy. Although FSGP also surpasses SGP—benefiting from richer feature fusion—it eventually saturates over prolonged operation. In contrast, FSGP-BGK retains its advantage over time, owing to the integration of historical observations via BGK, thereby yielding more robust and stable traversability estimates.

As the number of inducing points increases to 500, the performance gap between SGP and FSGP narrows, since the larger input feature space compensates for SGP’s simpler modeling. Nonetheless, FSGP-BGK still achieves about a 12\% improvement in mean accuracy compared to both baselines. This finding indicates that even with abundant inducing points, the BGK-based historical data fusion remains critical to preserving higher accuracy and stability, especially over extended trajectories.

\subsubsection{Comparison with Elevation Map}
We further validate our approach by comparing FSGP-BGK with the EM\cite{b11} method under identical conditions to those used for the baseline SGP. In this experiment, occlusion-induced gaps are excluded so that only observable regions are evaluated. Table~\ref{tab:comparison} reports the absolute differences in mean error and variance relative to the ground truth, along with the average runtime measured on the same hardware.

FSGP-BGK achieves a mean error of 0.1139 and a variance of 0.0333, compared to 0.1953 and 0.0422 for EM\cite{b11}, respectively. Moreover, its average runtime is only 33.84ms versus 107.85ms for EM\cite{b11}. These improvements can be attributed to the spatio-temporal continuity enforced by our BGK fusion and the efficient, feature-driven SGP framework. In contrast, the standard interpolation used in elevation maps yields coarser traversability estimates with higher variance and increased computational overhead. Consequently, FSGP-BGK produces a smoother, more accurate, and real-time traversability map that is well-suited for autonomous navigation.

\begin{table}[!htbp]
   \vspace{0.05cm}  
    \centering
    \caption{Comparison of FSGP-BGK and EM Methods. }
    \label{tab:comparison}
    \begin{tabular}{lccc}
    \toprule
    \textbf{Method} & \textbf{Mean} & \textbf{Variance} & \textbf{Avg.\ Runtime (ms)} \\
    \midrule
    FSGP-BGK & 0.1139 & 0.0333 & 33.84 \\
    EM       & 0.1953 & 0.0422 & 107.85 \\
    \bottomrule
    \end{tabular}
\end{table}

\subsubsection{Comparison of Historical Observational Data Fusion}
To demonstrate the superiority of our BGK-based approach within the FSGP framework, we conducted experiments comparing the direct accumulation of multi-frame point cloud data with our proposed BGK post-processing method. Under identical hardware conditions and in the same uneven terrain environment, we evaluated both approaches in terms of memory usage, GPU memory consumption, and processing time, as detailed in Table~\ref{tab:resource_comparison}. Our results indicate that while both methods exhibit comparable system memory usage (9.4\%), the BGK-based method substantially reduces GPU memory consumption (25\% versus 36\%) and processing time (29.16 ms versus 42.37 ms). These improvements are attributed to the effective integration of historical data via our BGK framework, which not only maintains high estimation accuracy but also significantly lowers computational overhead, thereby enhancing real-time performance.

\begin{table}[!htbp]
\vspace{-0.2cm}
\centering
\caption{Resource Usage and Performance Comparison}
\label{tab:resource_comparison}
\begin{tabular}{lcc}
\toprule
 & \textbf{FSGP-BGK} & \textbf{FSGP-Accumulation} \\ 
\midrule
\textbf{Memory Usage (\%)}      & 9.4   & 9.4   \\
\textbf{GPU Memory Usage (\%)}  & 25    & 36    \\
\textbf{Processing Time (ms)}   & 29.16 & 42.37 \\
\bottomrule
\end{tabular}
\end{table}

\subsection{Real-World Tests}

\begin{figure}[h]
   \vspace{0.05cm}  
  \centering
  \includegraphics[width=\columnwidth]{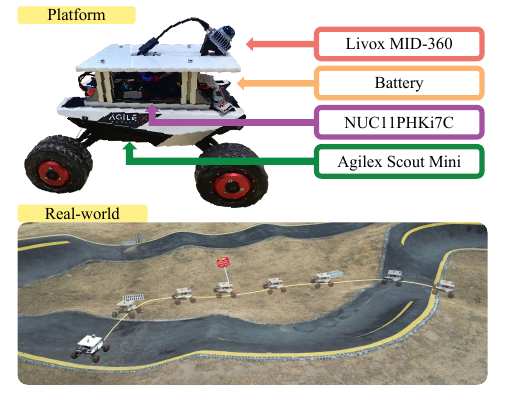}
  \caption{The physical platform used in our experiments (Agilex Scout Mini) is equipped with a DJI Livox MID-360 LiDAR. Two batteries are installed to power both the LiDAR and the NUC11PHKi7C. The figure also shows the real uneven terrain employed for our field experiments.}
  \label{fig:platform}
    \vspace{-0.3cm}   
\end{figure}

We deploy both FSGP-BGK and SGP on a differential-wheeled robot (as shown in Fig.~\ref{fig:platform}), which is equipped with an 11th Gen Intel\textsuperscript{\textregistered} Core\texttrademark{} i7 CPU, an NVIDIA\textsuperscript{\textregistered} RTX 2060 GPU, and a LiDAR sensor. As illustrated in Fig.~\ref{fig:platform}, this configuration enables real-time scanning and mapping of the surrounding environment, providing a robust testbed to evaluate the qualitative performance of our proposed approach.

\begin{figure}[h]
  \centering
  \includegraphics[width=\columnwidth]{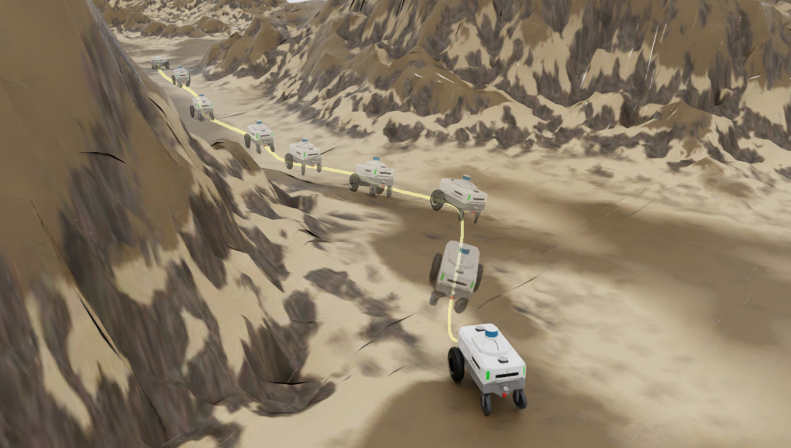}
  \caption{Smooth and continuous trajectory generation based on traversability maps in the Isaac Sim}
  \label{fig:carSim}
      \vspace{-0.5cm}   
\end{figure}

\paragraph{Autonomous Navigation}
We conducted real-world navigation experiments on uneven terrain, as shown in Fig.~\ref{fig:platform}, to verify the practical applicability of our traversability map on a physical robot. Fig.~\ref{fig:carSim} shows a smooth, continuous trajectory generated in Isaac Sim based on the traversability map produced by FSGP-BGK.


\begin{figure}[h]
   \vspace{0.05cm}  
  \centering
  \includegraphics[width=\columnwidth, trim={0.0cm 0.3cm 0.0cm 0.0cm}, clip]{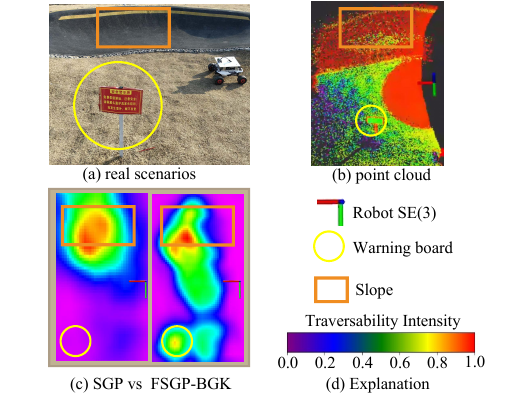}
  \caption{(a) Real-world uneven terrain (with small obstacles); (b) Scene obtained from point cloud extraction; (c) Comparison of traversability maps; (d) Traversability range from 0 to 1 (lower values indicate higher traversability).}
    \vspace{-0.5cm}   

  \label{fig:experiment}
  
\end{figure}

\paragraph{Real-World Map Perception Evaluation}
We further evaluate map perception in a real-world environment characterized by significantly undulating terrain, as shown in Fig.~\ref{fig:experiment}(a). FAST-LIO2 is employed to obtain high-fidelity point clouds and provide accurate localization, revealing small-scale obstacles in the scene (Fig.~\ref{fig:experiment}(b)). When the baseline SGP is run on the same hardware with identical induced points, resolution, and control pipelines, its resulting traversability map fails to capture these detailed information. In contrast, our FSGP-BGK algorithm, operating under the same conditions, accurately identifies these small obstacles at a frequency of 20\,Hz (Fig.~\ref{fig:experiment}(c)). Moreover, FSGP-BGK maintains a significantly lower computational overhead compared to SGP, thereby confirming its advantage in real-time perception and planning for challenging outdoor navigation tasks.

\section{Conclusion}
In this paper, we presented a global-map-free navigation framework that leverages a feature-based sparse Gaussian process to extract key geometric features from LiDAR point clouds. By integrating GPU-accelerated feature extraction with spatial-temporal Bayesian Gaussian Kernel inference, our method fuses real-time measurements with historical data to dynamically assess traversability in terms of slope, flatness, gradient, and uncertainty. Both simulation and real-world experiments confirm that the proposed approach significantly enhances estimation accuracy and computational efficiency compared to conventional methods, making it a robust solution for autonomous navigation in complex terrain.For future work, we will extend our approach to global-map-free autonomous exploration for multi-robot systems using GP and BGK.

\bibliography{references}

\end{document}